# Random Forests and VGG-NET: An Algorithm for the ISIC 2017 Skin Lesion Classification Challenge

Songtao Guo, Yixin Luo, and Yanzhi Song[*]

University of Science and Technology of China

## 1. Introduction

This manuscript briefly describes an algorithm developed for the ISIC 2017 Skin Lesion Classification Competition.

In this task, participants are asked to complete two independent binary image classification tasks that involve three unique diagnoses of skin lesions (melanoma, nevus, and seborrheic keratosis).

In the first binary classification task, participants are asked to distinguish between (a) melanoma and (b) nevus and seborrheic keratosis. In the second binary classification task, participants are asked to distinguish between (a) seborrheic keratosis and (b) nevus and melanoma.

The other phases of the competition are not considered.

Our proposed algorithm consists of three steps: preprocessing, classification using VGG-NET [2] and Random Forests [3], and calculation of a final score.

## 2. Algorithm

Our algorithm is inspired by [1], the overall procedure is shown in Figure 1.

### 2.1 Preprocessing

As shown in Figure 1, *step1* will find out *Lesion Area* in *Origin Image*. In this step, we have no need to delineate the boundaries of the lesion precisely. Areas with sufficient information is enough. In view of this, we have not designed a model specifically for this step, but used a model of our previous projects, mainly designed by convolutional neural network. The model is robust enough and does a good work here.

*Corresponding author, email:  songyz@mathu.cn.*

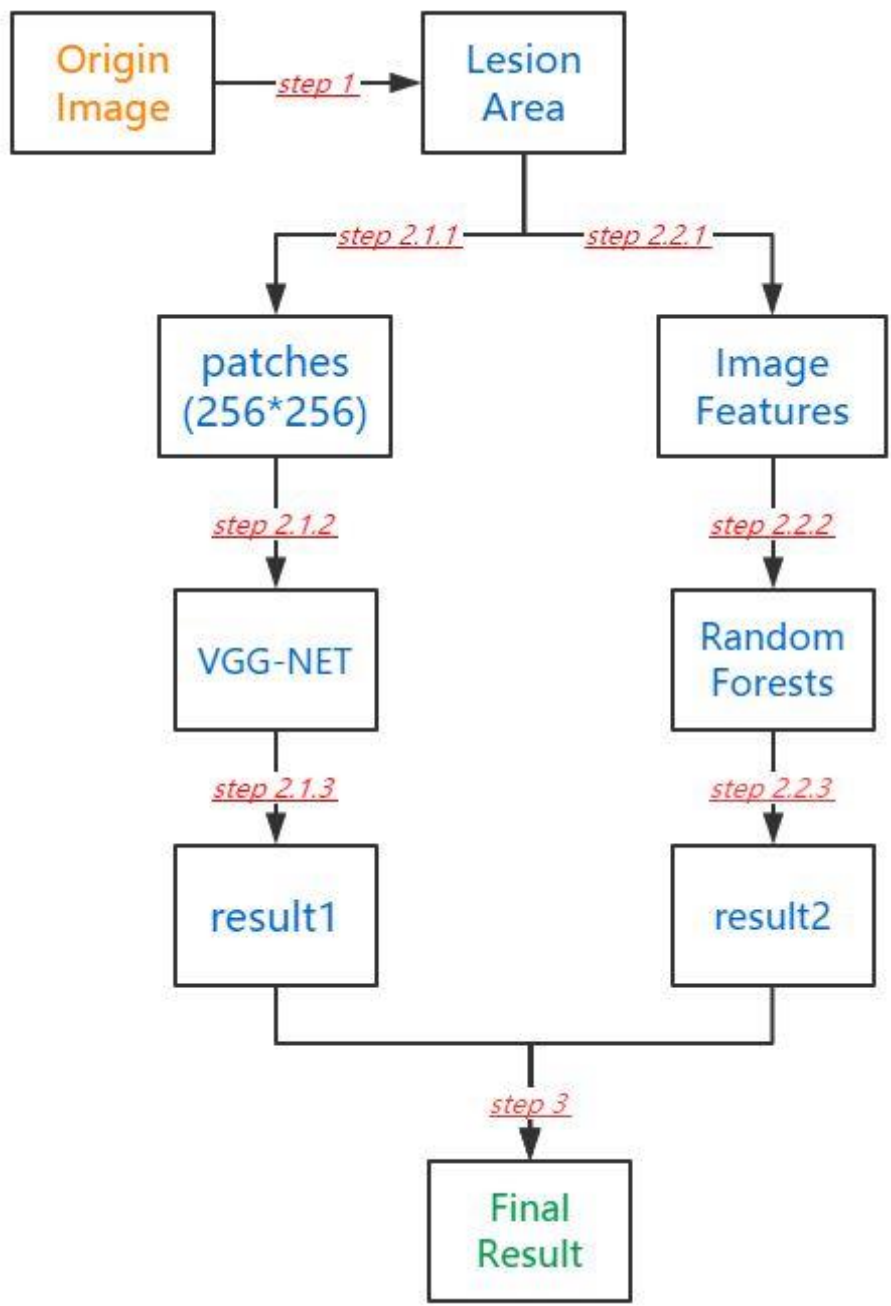

*2.2 Classification*

In this section, we adopt two methods to classify. One is *VGG-NET* and the other is *Random Forests*. Need to mention that we have extended the training set when we trained the model in order to get the classifier learn enough knowledge.

*VGG-NET*

Inputs of *VGG-NET* are patches obtained by cutting or adjusting the images we get from *preprocessing*. The size of patches is required by the network to be 256*256. After *VGG-NET*, softmax function is used to convert the output to probabilities of diseases. Multiple results of patches from same image, and we get *result1*.

*Random Forests*

In this part, we firstly extract image features, including *color histogram feature*, *color moment feature*, *Gray Level Co-occurrence Matrix*, *LBP feature* and *HOG feature* [4]. Then we combine the obtained features into a final feature vector. There is an adaptive mechanism to ensure the consistency of the final feature dimension. We get *result2* through *RandomForests*. Note that *result2* is also a probability vector.

## *2.3 Evaluation*

By combining *result1* and *result2* with different weights, we obtain a final probability vector. Obviously, the disease with a probability larger than 0.5 is the classification result.

## *3.Conclusion*

By combining *CNN* and *Random Forests*, our classifier takes into account color, texture, and geometry of the target image, and achieves very good performance on the test set.